\documentclass[preprint,review,12pt]{elsarticle}

\usepackage{lineno,hyperref}
\usepackage{amsmath,amssymb,amsfonts}
\usepackage{subcaption}
\usepackage{graphicx}
\usepackage{multirow}
\usepackage{booktabs}

\usepackage{algorithm}
\usepackage{algorithmic}

\modulolinenumbers[5]

\newcommand{\vect}[1]{\mathbf{#1}}

\journal{Biomedical Signal Processing and Control}

\begin{document}

\begin{frontmatter}

\title{An Explainable Fast Deep Neural Network for Emotion Recognition}

\author{Francesco Di Luzio}
\ead{francesco.diluzio@uniroma1.it}
\author{Antonello Rosato}
\ead{antonello.rosato@uniroma1.it}
\author{Massimo Panella\corref{cor1}}
\ead{massimo.panella@uniroma1.it}

\cortext[cor1]{Corresponding author}

\address{Department of Information Engineering, Electronics and Telecommunications,\\University of Rome ``La Sapienza'', Via Eudossiana 18, 00184 Rome, Italy.}

\begin{abstract}
In the context of artificial intelligence, the inherent human attribute of engaging in logical reasoning to facilitate decision-making is mirrored by the concept of explainability, which pertains to the ability of a model to provide a clear and interpretable account of how it arrived at a particular outcome.
This study explores explainability techniques for binary deep neural architectures in the framework of emotion classification through video analysis. We investigate the optimization of input features to binary classifiers for emotion recognition, with face landmarks detection using an improved version of the Integrated Gradients explainability method. 
The main contribution of this paper consists in the employment of an innovative explainable artificial intelligence algorithm to understand the crucial facial landmarks movements during emotional feeling, using this information also for improving the performances of deep learning-based emotion classifiers. 
By means of explainability, we can optimize the number and the position of the facial landmarks used as input features for facial emotion recognition, lowering the impact of noisy landmarks and thus increasing the accuracy of the developed models.                                
In order to test the effectiveness of the proposed approach, we considered a set of deep binary models for emotion classification trained initially with a complete set of facial landmarks, which are progressively reduced based on a suitable optimization procedure.

The obtained results prove the robustness of the proposed explainable approach in terms of understanding the relevance of the different facial points for the different emotions, also improving the classification accuracy and diminishing the computational cost.
\end{abstract}

\end{frontmatter}


\section{Introduction}
\label{sect:intro}
Deep Learning (DL) models have recently achieved admirable performance for classification and forecasting tasks, and nowadays they represent a crucial tool in a wide range of real-world applications (e.g. computer vision, document analysis, natural language processing, etc.) \cite{zhang2018survey, bai2021explainable, dong2021survey, ROSATO2021116852}. 
Acknowledged for their high predictive and classification accuracy, DL models are often identified with black-box methods that, given an input, offer little visibility into why specific data characteristics are selected over others for the generalization task.
Moreover, there is no explanation of how the training data is correlated with the represented choice of these features, or why specific pathways in the network are selected \cite{chakraborty2017interpretability}. 
For this reason, DL architectures are seen as limited models in terms of explainability and interpretability \cite{escalante2017design}. 
In addition to this, DL models are known to be resources and time-consuming methods, including in their architectures computationally expensive layers for classification and prediction purposes \cite{rosenfeld2019intriguing,zhang2020error,di2021blockwise,IANNELLI2022635}.  

As the fields in which Artificial Intelligence (AI) is committed are numerous, researchers, companies, and common users started to rely on its decision-making mechanisms; for this reason, the interpretability of their decisions and control over the inner processes of the different applications became, during the last years, a serious concern for several high-impact tasks \cite{choo2018visual}. 
The interpretation is the process of generating human-understandable explanations of why the different decisions are taken from DL models. 
For its inner nature, deep architectures hide the complex logic behind the decision-making process, making it difficult to reach such useful interpretations. 
For this reason, the employment of explainability for a better understanding of the inner comprehension of DL models is the key concept for a fair, safe, and understandable use of AI, in addition to a possible enabler for its deployment into everyday activity \cite{singh2020explainable}. 

The explainability and interpretability terms are often used in a commutable way in the literature; however, a distinction between the two concepts has been made in \cite{stano2020explainable}, where the interpretation refers to the mapping of abstract concepts like the output class into a domain example; explanation, on the other hand, refers on a set of domain features contributing to the decision-making process of the DL model.
The key role of explainability is the possibility given to the users to understand and reason about the inner mechanism of a deep architecture, leading to the model output. 
However, despite the numerous research works in this area, the progress remains quite limited in terms of the generalization capability of the implemented approaches\cite{lipton2018mythos}.  

In this paper, we are going to extend the methodology proposed in \cite{DILUZIO2023104418}; namely, we will describe and implement an improved version of the Integrated Gradients (IGs) algorithm developed on single binary classifiers for emotion recognition with landmarks detection \cite{sundararajan2017axiomatic}. 
In detail, we propose in this paper a new and more global version of this explainability technique, which examines the training mechanism of the implemented neural network on the entire training set and not on the single input sample.
To enhance the amount of information gained through using this method, we suggest some additional steps that are customized for the particular application presented in this research.

Thus, the novelty of this work is twofold and it is described in the following. 
From the explainability point of view, we exploit in-depth the inner machinery of the proposed method using IGs. 
The implemented approach, subsequently, allows us to discover which are the most important facial landmarks for each emotion characterizing the emotive feeling in human faces. 
In addition, from a practical point of view, we investigated the re-training and the optimization of several implemented binary deep neural networks (DNNs) using a limited number of facial landmarks as inputs to the model and basing the landmarks selection on the attributions of importance given to every facial point when optimizing the single binary emotion classifier.

We could finally evaluate and test the performance of the new proposed algorithm on real data, and the results will be compared with those of state-of-the-art methods.
In particular, the well-known CK+ dataset \cite{5543262} is employed for the experiments.

The results of the presented method are interesting from different perspectives.
First of all, the proposed explainable method is able to exploit the importance of every single input facial landmark for each emotion giving a numerical and graphical representation of the feature importance for the training of different binary classifiers. 
Moreover, from a computational point of view, the importance attribution of the facial landmarks is employed to optimize the number of input landmarks to the network, giving, as a result, an increase in the accuracy of the network lowering the computational burden of the training procedures.

The rest of the paper is organized as follows.
In Sect.~\ref{sect:related}, other investigations carried out on explainability in the emotion recognition field are presented. In Sect.~\ref{sect:material}, the overall explainable AI method implemented is presented. 
In Sect.~\ref{sect:results}, the results of the proposed methodologies on a well-known dataset are presented and they are successively discussed in Sect.~\ref{sect:discussion}. 
Finally, in Sect.~\ref{sect:conclusions} the concluding remarks are drawn.

\section{Related Work}
\label{sect:related}
Automatic emotion recognition has been a fascinating and extensively researched topic for over 40 years. 
Recent studies have shown that emotion recognition systems have the potential to be used in various applications, such as public safety, people surveillance, and human-computer interaction \cite{yang2011facial}. 
The objective of this section is to describe the innovative use of explainability methods in recent research studies to enhance the performance and interpretability of DNNs for emotion recognition. 
This literature review will examine the most recent and innovative research studies in the field of automatic emotion recognition that have employed explainability methods.
In detail, this paper will be centered on explainability with landmarks detection, but given the innovative aspect of this dissertation, the description will be enlarged including methods working with different streams of input data, from text analysis to visual samples.

Recently, different explainability approaches have been investigated for emotion recognition tasks. 
In \cite{kumar2022bert}, an explainable DL model is proposed for multi-class text emotion recognition. 
The authors developed a novel explainability technique for the training of an innovative prediction procedure. 
The proposed architecture is composed of different modules, the classification is done through the employment of convolutional layers and bidirectional Long Short-Term Memory (LSTM) layers. 
Explainability is used to describe the training and predictions of the proposed system analyzing the inter and intra-cluster distances, where the clusters are the projection of the emotion embeddings on a hyperplane by using, as illustrated in the following, a scale-based filtering approach \cite{MASCIOLI20001001}. 
The results achieved with this method are promising and ensure the applicability of the model to real-world applications and diverse texts. 
This text emotion recognition system for emotional classification can in fact classify sentiment classes with high precision and can be used for various textual data types. 

In spite of this, the proposed approach is applicable only to text data and it is only able to distinguish between positive, neutral and negative emotions, without the possibility to discern the single sentiments. 
Another interesting work in terms of explainable AI for emotion recognition is presented in \cite{kumar2021towards}. 
In this paper, a multimodal speech-emotion recognition system has been developed with an explainability framework to describe the prediction process of the network. 
The neural architecture is based on Gated Recurrent Unit (GRU) and pre-trained with a bidirectional encoder; for the prediction of the final emotion class, the layers are successively concatenated. 
The training of the layers and the prediction of the network have been explained through emotion embedding plots and analyzing the intersection matrices for various emotion classes' embeddings.  
In this case, the presented framework is interesting and reaches a high accuracy of the classification. 
On the other hand, the multimodal aspect of the method could be expanded with the incorporation of more modalities such as videos and images, which is what is properly studied in the present work.

More related to images, and facial interpretation, is the paper proposed in \cite{kandeel2021explainable}, where a Convolutional Neural Network CNN-based model is developed for driver emotion recognition, using the output of the network to expect his behavior during the drive. 
The model has been evaluated on four different datasets and some experiments have been conducted on a real driving environment with good results. 
Two different interpretability techniques have been used to understand the model behavior: the saliency map, which is one of the most known methods, which is used to clarify the importance of some areas in an input image; and Grad-CAM, another map that specifies the neuron's important values for each specified decision of interest. 
An interesting improvement of this work could be the analysis of the driver's distraction in real-time, which could lead to an extension of the applicability domains of the approach.

Another work using images of faces as input is presented in \cite{deramgozin2021hybrid}. 
In this investigation, a hybrid explainable artificial intelligence framework composed of a functional and an explainable block has been implemented for facial expression classification. 
Also in this case, the framework is based on a 6-layer convolutional neural network. 
This model has been backed by a layer comprising a facial action unit extraction module whose outputs are used for the interpretation of the obtained output. 
This module is based on an autoencoder that uses the pre-trained Resnet-50 as an encoder to extract the action units from the input image. 
A multilayer perceptron is added at the output of the extractor to reinforce the functional pipeline in terms of classification accuracy. 
This work is well-presented and reaches good accuracy, it would be even more powerful if used with an improved system of the facial action unit extraction module implemented with state-of-the-art neural methods.

Also in \cite{cesarelli2022emotion} an explainable DL algorithm for emotion recognition from human faces is built. 
In this case, the classification is performed between three basic emotions: happiness, neutrality, and sadness. The proposed method, which takes input images, is able to show on the facial images the areas that are symptomatic of a certain emotion. 
$1500$ samples have been used to train the proposed promising explainable emotion recognition method.
This is an inspiring work, presenting really interesting results and having some similarities with our proposed solution; however, the possibility of discerning between only three classes is quite limiting for several different practical applications. 

Given this literature review, the aim of the proposed paper is to learn from the proposed approaches incorporating the interesting aspects presented above and expand the domains and the applicability of explainability methods. 
The goal of the proposed approach is in fact to present a new global explainable framework outstanding the state-of-the-art for generalization capability and for completeness of applicability to the entire set of the six typical emotions.

\section{Material and Methods}
\label{sect:material}
Given the importance of this topic, highlighted also by the preceding literature review, in the following we present the investigation of explainable binary DNNs for emotion recognition.
The starting point of this paper is presented by the investigation presented in \cite{DILUZIO2023104418}, in which deep binary classifiers are implemented to distinguish a single emotion from the others. 
In this paper, we present the problem of attributing the classification of these deep binary classifiers to their input features, which in this case are 468 facial landmarks coordinates. 
Moreover, we use these attributions vectors to optimize the number and the selection of the input features leading to better performance of the implemented models and for a complete understanding of the facial point importance in the classification of the single emotions. 

Let $F : \mathbb{R}^n \rightarrow{} [0,1] $ represent a binary classifier and ${\mathbf{x} = (x_1,\dots,x_n) \in \mathbb{R}^n}$ a general input; an attribution of the prediction of input $\mathbf{x}$, relative to a \emph{baseline} input $\mathbf{x}^{'}$, is a vector $\mathbf{a}=A_F(\mathbf{x},\mathbf{x}^{'})$ where $\mathbf{a}=(a_1,...,a_n) \in \mathbb{R}^n$ and $a_i$ is the contribution of the $i$-th feature of input $\mathbf{x}$ to its prediction. 
Establishing a baseline facilitates the interpretation of the methodology, such that the attributions are independent of the baseline, and thus, the output can be attributed to the specific input features of the samples under examination.

\subsection{Integrated Gradients}
Given that the concept of baseline is fundamental to describe the functioning of the deep classifier with respect to the input we propose an explainability method for the attribution of importance of the input features based on IGs, which define an explainability technique presented in \cite{sundararajan2017axiomatic} based on the concept of baseline. 
The baseline $\mathbf{x}^{'}$ is represented by a sample that has the same dimensions as the DNN input and it is chosen in a way that the classification probability at the baseline is near zero. 
For example, the general choice when applying IGs to a DNN trained with images is the completely black image.
IGs are defined ``as the path integral of the gradients along the straight-line path from the baseline $\mathbf{x}^{'}$ to the input $\mathbf{x}$'' \cite{sundararajan2017axiomatic}. 
Hence, the IG along a dimension is the gradient of the objective function $F(\mathbf{x})$ along that dimension. 
In our implementation, the function $F$ will be represented by the different binary DNNs. 

The IGs technique is an interesting method for several different reasons: first, it satisfies \textit{sensitivity}, meaning that for every input and baseline differing in one feature and in the predictions, the different feature is given a non-zero attribution. 
Moreover, this method satisfies also \textit{completeness}, which testifies that the attributions add up to the difference between the output of the objective function at the input under exam and the baseline.

The IG for the $i$-th feature of input $\mathbf{x}$ is defined as:
\begin{equation}
    \mathrm{IG}_{i}(\mathbf{x}) = (x_{i} - x^{'}_{i})\cdot{\int_{0}^1\frac{\partial F\left[\mathbf{x}^{'}+\alpha(\mathbf{x} - \mathbf{x}^{'})\right]}{\partial x_i}~{d\alpha}}\,,
    \label{eq:int_gradients_1}
\end{equation}
where $\alpha$ is an interpolation constant employed for a perturbation of the features. From a practical point of view, it is not always numerically possible to solve the integral presented in \ref{eq:int_gradients_1} and it can be a computationally costly operation; hence, the numerical approximation can be computed as:
\begin{equation}
    \widetilde{\mathrm{IG}}_{i}(\mathbf{x})=\frac{(x_{i}-x^{'}_{i})}{m}\cdot{\sum_{k=1}^{m}\frac{\partial F\left[\mathbf{x}^{'} + \frac{k}{m}(\mathbf{x} - \mathbf{x}^{'})\right]}{\partial x_{i}}}\,,
    \label{eq:int_gradients_2}
\end{equation}
where $k$ is the scaled feature perturbation constant and $m$ is the number of steps in the Riemann sum approximation of the integral. 
In particular, in this discretized implementation of IGs the scaled feature perturbation constant $k$ singularly describes each of the $m$ steps, $k=1\dots m$, of the tensor representing the interpolated samples. Increasing the number $m$ of interpolation steps will increase the number of possible values for $k$, making \eqref{eq:int_gradients_2} a more accurate approximation of \eqref{eq:int_gradients_1}.

Referring to the term $(x_i - x^{'}_{i})$, this is a necessary term used to scale the IGs and keep them in terms of the original sample. 
This represents the path from the baseline sample to the input one. 
Given that IG consists of the integration in a straight line, this sum ends up being roughly equivalent to the integral term of the derivative of the interpolated sample function with respect to $\alpha$ with a high number of steps. 
Hence, the first thing to do is to generate a linear interpolation between the baseline and the original input. 
From a practical point of view, the interpolated samples are small steps in the feature space between the baseline $\mathbf{x}^{'}$ and the input $\mathbf{x}$. 
Then, the gradients are computed, to measure the path between changes in a feature and changes in the model's prediction. 

Given the construction of the method, IGs present some limitations: 
\begin{itemize}
    \item  IGs algorithm provides features importance on individual examples, but it does not reveal particular information across the entire dataset;
    \item the method provides individual feature relevance, but it does not explain feature interactions and combinations;
    \item even if IGs can be used to help understand how the network works, if the features highlighted as more informative are not the ones matching a logical intuition, there is no explanation for why this phenomenon takes place.  
\end{itemize}

Given those limitations, which can be quite significant in terms of the generalization capability of the approach, we propose in Subsection\ref{sect:proposed_approach} some additional steps that are meant to overcome these limitations intrinsic to IGs.
In particular, increasing the globalization capability of the method over the whole dataset enables the use of the proposed explainable technique for the optimization of the number of input landmarks to the DNN. 

\subsection{Data Environment}
\label{sect:data_env}
Given the preceding summary on IGs technique, for a better understanding of the proposed methodology and for a good comprehension of the experimental activity, we will describe the data manipulated and used in this work.
We employed for training, validating, and testing one of the most known and important datasets in the field of emotion recognition, the Extended Cohn-Kanade dataset (CK+) \cite{5543262}. 
The CK+ contains videos of emotional feelings, it contains sequences showing the shift from neutral to peak expression, and it is composed of videos representing 7 different emotions, including the \  16 classical emotions with the addition of the contempt emotion. 
The dataset is composed of videos of students with age between 18 and 30 years old. 
$65\%$ of the subjects in the dataset are women, $15\%$ of the data samples are African-Americans, and $3\%$ of the subjects are Asian and South American. 
All the videos in the dataset have $640 \times 480$ pixel resolution at $30$ fps, with 8-bit grayscale images. 
Despite its limitations, such as the narrow age range and lack of representation of various ethnic groups, this dataset is commonly used to study DL models for recognizing emotions. 
It is useful for comparing a model's performance to the current leading models and provides a reliable performance metric. 
To increase the dataset for evaluation purposes, we also performed data augmentation by flipping each frame of each video vertically. 
This ensures the model's dependability in diverse conditions.
However, given the low number of instances of the dataset labeled with the `contempt' label, this emotion is not analyzed in the proposed study, and data samples with that label have been removed from the employed dataset. 
Each frame in each video undergoes the preprocessing procedure, consisting of:
\begin{itemize}
\item face cropping from each video;
\item resizing of the square face box image to a $300 \times 300 \times 1$ image, as one-channel (grayscale) image;
\item detection of the 468 face landmarks, depicting the entire face.
\end{itemize}
The 468 landmarks plane coordinates of each frame of each video represent the input features of the neural architecture. 
In detail, the $i$-th sample of the dataset is represented by a $P\times 2 \times L$ tensor $\vect{V}^i$, where $P$ is the number of video frames and $L$ is the number of extracted landmarks.  
One of the most important contributions of the presented approach consists in the employment of an improved and suitable fashion of the explainable AI technique of IGs for the optimization of $L$.

\subsection{Proposed Approach}
\label{sect:proposed_approach}
The proposed explainability and optimization method presented in this paper is based on the application of the IGs to several binary DNNs (i.e. six, one for each typical emotion) for binary emotion recognition with landmarks detection. 
The network architecture is the same as the deep model presented in \cite{DILUZIO2023104418}. 
It consists of stacked Convolutional and LSTM layers employed to extract spatio-temporal features from the videos given in input to the network. 
These extracted features are subsequently passed to fully-connected layers with non-linear activation functions for the final binary classification. 
To deal with the high-computational burden typical of the training of DNNs, two different fashions of the model are proposed: R-EMO (Randomized Emotion classifier) in which the convolutional layer is randomized and T-EMO (Trained Emotion classifier), the fully trained version of the network. 

\begin{figure}
	\centering
		\includegraphics[width=0.75\columnwidth]{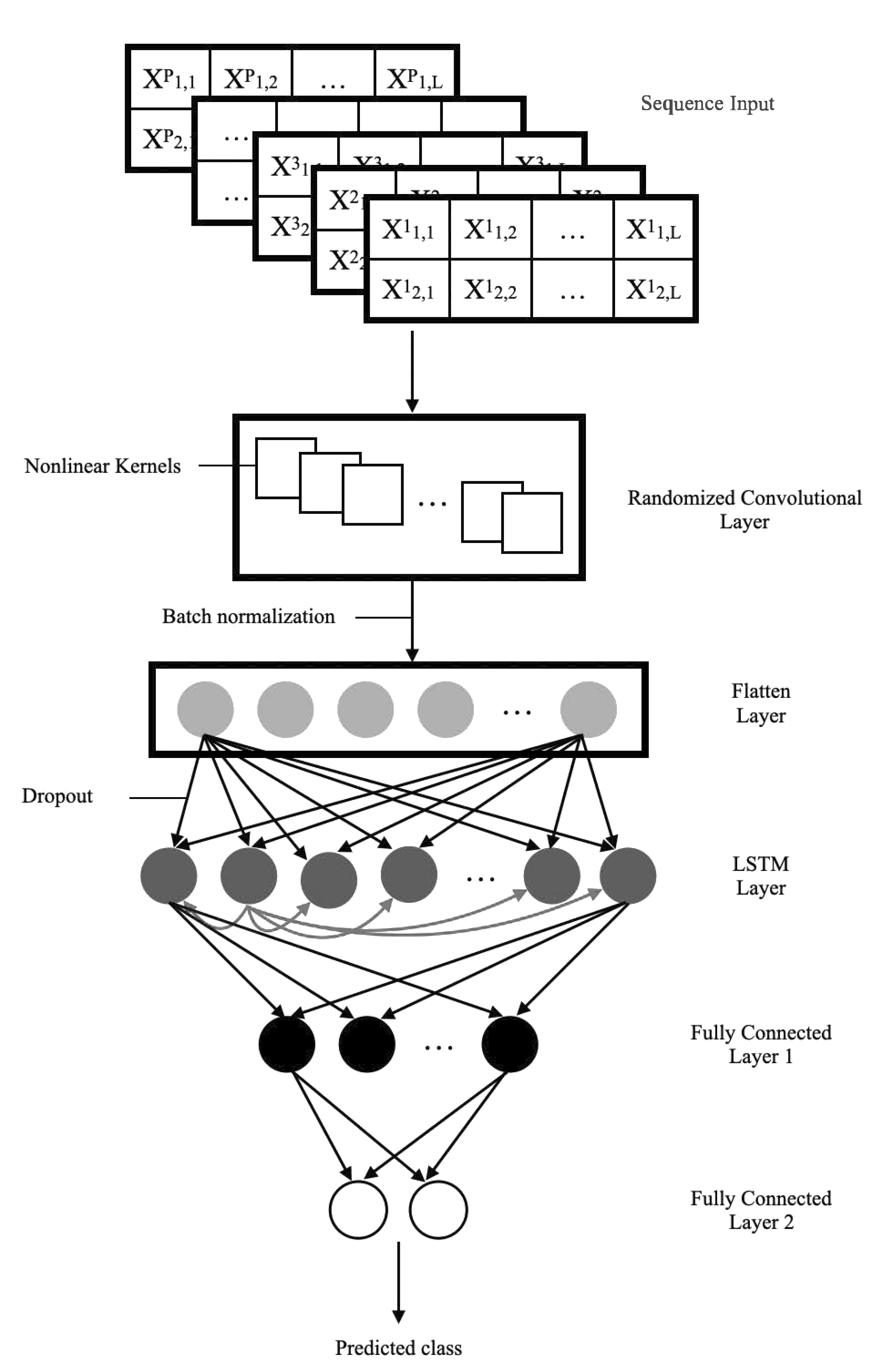}
	\caption{Graphical representation of the proposed DNN architecture that is used for emotion recognition/classification. The generic element $X^{i}_{j,n}$ in each input slice, which is associated with a video frame, represents the horizontal/vertical spatial coordinate (${j=1,2}$) of the $n$-th landmark (${n=1\dots L}$) within the $i$-th video frame (${i=1\dots P}$).}
	\label{fig:net_struct}
\end{figure}

Moreover, applying explainability to randomized networks is something particularly interesting, because it investigates the typical trade-off between the accuracy of the classification and efficiency of the training procedure from a different perspective. In fact, what is expected is a possible gain in accuracy and efficiency given by the explainable procedure and its consequent optimization of the input parameters which can further balance the loss in accuracy of the classification, result of the randomization of some layers in the network. 
A graphic representation of the proposed network is presented in Fig.~\ref{fig:net_struct}. More in detail, the network is composed of:
\begin{itemize}
    \item a one-channel 2D-CNN layer which receives a frame of the video in input and applies various different convolutional filters associated with nonlinear kernels with Rectified Linear Unit (ReLU) as activation function. 
    The result of this layer is represented by different feature maps. 
    In the case of the proposed R-EMO network, the entire set of weights of the kernels is selected randomly and never changed during the network optimization. 
    Instead, in the fully trained version of the network T-EMO, the filters' weights are estimated by the training procedure;
    \item a batch normalization layer for normalizing the inputs from the convolutional layer using the means and standard deviations of the batches of inputs seen during training;
    \item a flatten layer used to collapse the tensor associated with the normalized feature maps;
    \item an LSTM layer which is the only recurrent layer of the proposed architecture;
    \item a dropout layer employed after the recurrent layer in order to prevent overfitting;
    \item a fully connected layer (denoted as FC1) connecting the hidden state of the preceding LSTM layer (possibly after the drop-out operation) to the succeeding final layer for classification;
    \item another fully connected layer (denoted as FC2) with two neurons for the final classification.
\end{itemize}
%
The particular architecture described so far presents some interesting aspects when related to its final application and methodology. 
First of all, the proposed model is capable of tracking the spatio-temporal correlation among different landmarks of the several frames in the input videos, by using a combination of CNN layers and LSTM layers. 
The importance of this aspect is greatly highlighted by the application under investigation.
In fact, the videos show the variations of human faces during the emotional feeling and hence, with this combination, it is possible to elaborate the relative spatial variation of the landmark coordinates across different frames. 
Moreover, the sole employment of landmarks coordinates in each frame as input to the network allows the optimization of the model weights by a relatively light computational procedure, thus with the possibility to develop a real-time classification software.
After the CNN and the LSTM layers, the Fully Connected layers are employed to analyze a suited projection of the extracted features using the nonlinear activation functions of such layers, just as a common practice when building deep learning models for classification.

When computing IGs through \eqref{eq:int_gradients_2}, for the specific application presented in this paper, the function $F$ is represented by the binary DNNs themselves. 
Once trained, the model is in fact a learned function describing one of the possible mappings between the input feature space and the output space defined by the different classes (i.e., the six different emotions). The expected input for the model is a dense 5D tensor with shape $N\times P\times C\times L\times 1$ where:
\begin{itemize}
    \item $N$ is the number of video samples in the dataset;
    \item $P$ is the number of video frames employed for each sample;
    \item $C$ is the number of spatial coordinates (i.e., $C=2$);
    \item $L$ is the number of landmarks that is the parameter to be optimized by using the proposed explainable AI technique.
\end{itemize}
The latest tensor dimension is 1 as we are using as input to the DNN an equivalent one-channel (grayscale) sample. 

\begin{figure}
	\centering
		\includegraphics[width=0.8\columnwidth]{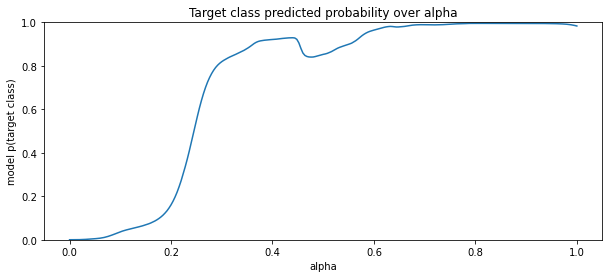}
	\caption{Graphical representation of the target class probability for the emotion surprise over alpha when using a sadness-labeled video as the baseline. In particular, it can be seen that when $\alpha$ is $0$, the network predicts the interpolation class as \textit{non-surprise}. When $\alpha$ increases, the model confidence in the \textit{surprise} classification increases as well saturating after $\alpha = 0.3$ }
	\label{fig:over_alpha}
\end{figure}

The gradient is used to reveal which landmarks coordinates of the frames have the strongest effect on the model's predicted class probabilities given that the gradients are describing the local changes in the model's output. 
From a visual point of view, the plot in Fig.~\ref{fig:over_alpha} shows how the model's confidence in the prediction varies across alphas which can be interpreted as the small steps in the feature space between the baseline and the input. 
For this graphic representation, the single sample is labeled as \textit{surprise}, and the baseline represent one of the \textit{happiness} emotional feeling video. 
After the IGs are computed, their absolute values across the frames and across the coordinates are summed to produce an attribution mask. This attribution vector shows the information brought by each landmark throughout the entire video. 

To deal with a subset of IGs limitations we implemented a novel algorithm, in particular, trying to give a more global fashion to the proposed approach. 
In fact, to obtain a greater amount of information from the utilization of this method, we propose some additive steps, tailored for the specific application proposed in this study:
\begin{itemize}
\item  for each emotion, the selection of a video representing a typical behavior of that label is performed;
\item Then, for each label, IGs are computed using as $F$ the binary network related to that emotion, computing IGs for each input sample in the training set and employing as a baseline one of the five typical samples;
\item the whole process is repeated using the five different baselines selected as typical representations of the complementary emotions (i.e. the five different labels with respect to the examined input sample); 
\item after the computation of the IGs for all the combinations, for each emotion, the attributions obtained with each sample are firstly averaged along the samples channel, and successively along the five respective baselines channel for each emotion, obtaining a final attribution vector. 
\end{itemize}

Following this procedure, we obtained, for each label, an average value of the attribution mask reflecting the importance of every single landmark for the classification of that specific emotion, being thus able to organize the $468$ input features from the most important to the least informative. 
This is one of the main contributions of this paper, from the explainable AI perspective, it allows the awareness of the importance of every single landmark for the detection of each emotion in a video sequence. 
From a practical point of view, using $468$ landmarks as input features can be useful for different reasons: first of all, it permits obtaining a complete depiction of the human faces, moreover, it allows capturing the facial micro-movements during the emotional feeling, being able to classify emotions with higher accuracy.

Using such a large number of input features could determine that some of them are not really helpful to the final classification and thus, adding noise to the overall classification method, diminishing the interpretability of the results. 
For this reason, it seems interesting to try different combinations of input features to the binary DNN before training, looking for an improvement in the classification performance basing the features selection on the IGs method proposed herein.

\section{Experimental Results}
\label{sect:results}
In this section, we present the numerical results obtained with the novel proposed methodology and the impact of explainability on the implemented method. 
In particular, we applied the proposed methodology to two different architectures (i.e. T-EMO and R-EMO) to demonstrate the improvement of the performance of the networks when deleting from the input features the non-informative landmarks. 
In addition, we were able to show from a graphical perspective how the selection of the most important facial points matches human intuition, reinforcing the solidity of the implemented models and the proposed explainable approach. 
Hence, with the aim of applying IGs to the six binary classifiers using \eqref{eq:int_gradients_2}, we employed as estimator $F$ the binary networks trained with the entire set of $468$ landmarks. Doing this, with the proposed method we obtained the attribution of the importance of each facial landmark for each emotion. 

\begin{figure*}
	\centering
		\includegraphics[width=0.75\columnwidth]{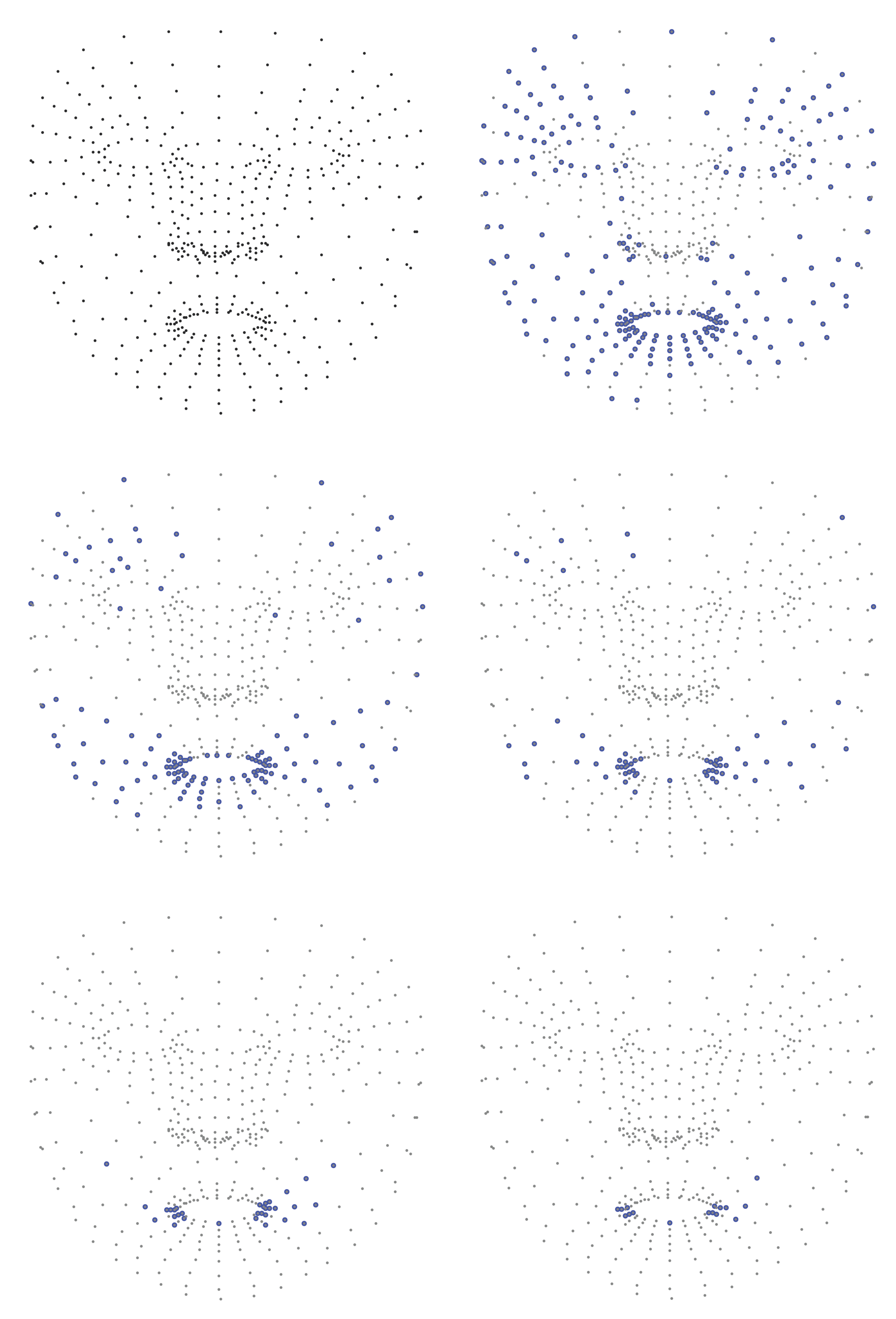}
	\caption{Graphical representation of the landmarks' relevance attribution, projected into a 2-D space. 
 The top left picture represents the 468 landmarks employed. 
 The other images show the most important landmarks selected over the entire set with a blue circle. 
 We have from the top right to the right down picture the different combinations of the most important landmarks: $234$. $128$, $64$, $32$, $16$.}
	\label{fig:surprise}
\end{figure*}

After obtaining these attributions related to each input feature, we performed six different grid search procedures for each binary network, for the optimization of the hyperparameters, changing for each attempt the number of input landmarks to a lower value: $234$, $128$, $64$ ,$32$ and $16$, respectively.
A graphical representation of the obtained importance attribution can be seen in Fig.~\ref{fig:surprise}.
The entire set of neural networks proposed in this paper is trained using the ADAM algorithm \cite{kingma2014adam} with a learning rate fixed at $10^{-4}$ and mini-batch size equal to $16$, selected after the first experimental activity. The grid search procedure is implemented on the training data for each of the experimental settings. 
This procedure has been followed in order to avoid overfitting issues. 
Given the available computational resources, the grid search procedures have been performed only to optimize the hyperparameters of the fully trained version of each binary network. 
From a theoretical point of view, it seems fair to evaluate the discrepancies in the performance obtained by the fully trained architectures and the randomized ones when the structure of the models is the same, in order to analyze the possible advantages and the potential losses due to the randomization of certain parts of the neural architectures. The different parameters obtained with the mentioned procedure are listed from Table~\ref{tab:net_struct_234} to 
Table~\ref{tab:net_struct_16}. 

\begin{table}[!ht]
	\caption{Optimized hyperparameters for T-EMO network with $234$ input landmarks.}
	\vspace{-12pt}
	\label{tab:net_struct_234}
	\begin{center}
	\footnotesize
	\begin{tabular}{l@{\quad\quad}c@{\quad}c@{\quad}c}
	\toprule
	\multirow{2}{*}{Emotion} & {Convolutional} &    {LSTM}     & {FC1}\\[-2pt]
	                         & {Filters ($F$)} & {Units ($Q$)} &  {Neurons ($R$)}\\           
	\midrule
	Anger       & $25$ & $31$ & $54$  \\ [2pt]
	Disgust      & $35$ & $50$ & $63$\\ [2pt]
	Fear     & $39$ & $51$ & $34$\\ [2pt]
	Happiness      & $50$ & $58$ & $63$\\[2pt]
	Sadness      & $24$ & $68$ & $54$ \\[2pt] 
  Surprise     & $30$ & $58$ & $63$\\[2pt]
	\bottomrule
\end{tabular}
\end{center}
\end{table}

\begin{table}[!ht]
	\caption{Optimized hyperparameters for T-EMO network with $128$ input landmarks.}
	\vspace{-12pt}
	\label{tab:net_struct_128}
	\begin{center}
	\footnotesize
	\begin{tabular}{l@{\quad\quad}c@{\quad}c@{\quad}c}
	\toprule
	\multirow{2}{*}{Emotion} & {Convolutional} &    {LSTM}     & {FC1}\\[-2pt]
	                         & {Filters ($F$)} & {Units ($Q$)} &  {Neurons ($R$)}\\           
	\midrule
	Anger       & $25$ & $31$ & $54$  \\ [2pt]
	Disgust      & $39$ & $30$ & $63$\\ [2pt]
	Fear     & $39$ & $51$ & $34$\\ [2pt]
	Happiness      & $39$ & $68$ & $72$\\[2pt]
	Sadness      & $39$ & $58$ & $74$ \\[2pt] 
  Surprise     & $40$ & $68$ & $73$\\[2pt]
	\bottomrule
\end{tabular}
\end{center}
\end{table}

\begin{table}[!ht]
	\caption{Optimized hyperparameters for T-EMO network with $64$ input landmarks.}
	\vspace{-12pt}
	\label{tab:net_struct_64}
	\begin{center}
	\footnotesize
	\begin{tabular}{l@{\quad\quad}c@{\quad}c@{\quad}c}
	\toprule
	\multirow{2}{*}{Emotion} & {Convolutional} &    {LSTM}     & {FC1}\\[-2pt]
	                         & {Filters ($F$)} & {Units ($Q$)} &  {Neurons ($R$)}\\           
	\midrule
	Anger       & $25$ & $31$ & $54$  \\ [2pt]
	Disgust      & $39$ & $30$ & $33$\\ [2pt]
	Fear     & $39$ & $51$ & $34$\\ [2pt]
	Happiness      & $16$ & $58$ & $52$\\[2pt]
	Sadness      & $39$ & $68$ & $64$ \\[2pt] 
  Surprise     & $40$ & $58$ & $63$\\[2pt]
	\bottomrule
\end{tabular}
\end{center}
\end{table}

\begin{table}[!ht]
	\caption{Optimized hyperparameters for T-EMO network with $32$ input landmarks.}
	\vspace{-12pt}
	\label{tab:net_struct_32}
	\begin{center}
	\footnotesize
	\begin{tabular}{l@{\quad\quad}c@{\quad}c@{\quad}c}
	\toprule
	\multirow{2}{*}{Emotion} & {Convolutional} &    {LSTM}     & {FC1}\\[-2pt]
	                         & {Filters ($F$)} & {Units ($Q$)} &  {Neurons ($R$)}\\           
	\midrule
	Anger       & $25$ & $31$ & $54$  \\ [2pt]
	Disgust      & $39$ & $50$ & $63$\\ [2pt]
	Fear     & $39$ & $61$ & $64$\\ [2pt]
	Happiness      & $16$ & $58$ & $52$\\[2pt]
	Sadness      & $39$ & $58$ & $74$ \\[2pt] 
  Surprise     & $30$ & $58$ & $63$\\[2pt]
	\bottomrule
\end{tabular}
\end{center}
\end{table}

\begin{table}[!ht]
	\caption{Optimized hyperparameters for T-EMO network with $16$ input landmarks.}
	\vspace{-12pt}
	\label{tab:net_struct_16}
	\begin{center}
	\footnotesize
	\begin{tabular}{l@{\quad\quad}c@{\quad}c@{\quad}c}
	\toprule
	\multirow{2}{*}{Emotion} & {Convolutional} &    {LSTM}     & {FC1}\\[-2pt]
	                         & {Filters ($F$)} & {Units ($Q$)} &  {Neurons ($R$)}\\           
	\midrule
	Anger       & $25$ & $31$ & $54$  \\ [2pt]
	Disgust      & $39$ & $50$ & $53$\\ [2pt]
	Fear     & $24$ & $68$ & $64$\\ [2pt]
	Happiness      & $16$ & $58$ & $52$\\[2pt]
	Sadness      & $39$ & $58$ & $64$ \\[2pt] 
  Surprise     & $40$ & $58$ & $63$\\[2pt]
	\bottomrule
\end{tabular}
\end{center}
\end{table}

The dataset employed for the experiments is the CK+ dataset, already described in Subsection \ref{sect:data_env}. 
A total of 487 video samples were utilized for training the networks, and 87 samples were used for testing various network architectures. 
For each experiment, 10\% of the training set was reserved for validation purposes. 
When training the binary classifiers, it should be noted that the training set had an imbalanced distribution, as it contained samples from six different emotions, with only one emotion being labeled with a `true' logical label (binary value 1) and the samples from the remaining emotions being labeled with a `false' logical label (binary value 0). 

Once found the optimal hyperparameters, for each combination of model parameters and number of features (i.e. facial landmarks), six binary classifiers are modeled and trained, with each of these models acting as a binary categorizer for every single emotion.

The model performance is evaluated by means of classification accuracy, intended as the ratio of well-classified samples over the number of total samples in the examined dataset. The binary cross entropy  $\operatorname{BC}$ is employed as loss function and is defined by \eqref{eq:bc}.
\begin{equation}
    \operatorname{BC} = -\frac{1}{S}\sum_{i=1}^S\big[y_i\log(\hat{y}_i)+(1-y_i)\log(1-\hat{y}_i)\big]\,,
    \label{eq:bc}
\end{equation}
where $S$ is the cardinality of the dataset, $y_i$ is the true binary label of the $i$-th sample and $\hat{y}_i$ is the probability estimated by the adopted model that the $i$-th sample represents the considered emotion.
Given that the training outcome and the consequent accuracy level on each test depend upon the random initialization of the model weights, 10 different runs of each algorithm are performed with 10 different seeds and with a different set of hyperparameters, varying also the training/validation/test set partition. 
Hence, the choice of the optimal hyperparameters and the coherent binary classification performance are based on the mean and on the standard deviation of the accuracy over the different runs of the proposed algorithm.
The binary classification performance of the different networks trained with the most informative landmarks is summarized in Table~\ref{tab:net_res_T} and Table~\ref{tab:net_res_R}.

\begin{table}[!ht]
	\caption{Classification accuracy for binary T-EMO network.}
	\vspace{-6pt}
	\label{tab:net_res_T}
 \addtolength{\leftskip} {-2cm}
    \addtolength{\rightskip}{-29cm}
        \centering
	\footnotesize
	\makebox[1.0\textwidth]{\begin{tabular}
 {l@{\quad}c@{\quad}c@{\quad}c@{\quad}c@{\quad}c@{\quad}c}
	\toprule
     &  468 & 234 & 128 & 64 & 32 & 16  \\
	\midrule
	Anger       & $0.924\pm0.016$ & \boldmath{$0.926\pm0.037$} & $0.905\pm0.037$ & $0.903\pm0.034$ &  $0.924\pm0.030$ &  $0.891\pm0.041$ \\ [2pt]
	Disgust     & $0.934\pm0.019$ & $0.964\pm0.027$ & \boldmath{$0.964\pm0.019$} & $0.951\pm0.034$ &  $0.948\pm0.040$& $0.925\pm0.045$\\ [2pt]
	Fear        & $0.948\pm0.024$ & $0.971\pm0.023$ & \boldmath{$0.974\pm0.020$} &  $0.964\pm0.026$ &  $0.950\pm0.026$ & $0.919 \pm 0.041$\\ [2pt]
	Happiness   & $0.986\pm0.008$ & \boldmath{$0.988\pm0.011$} & $0.982\pm0.014$ & $0.979\pm0.018$ & $0.981 \pm 0.015$& $0.978 \pm 0.012$ \\ [2pt]
	Sadness     & $0.920\pm0.026$ & $0.905\pm0.052$ & $0.916\pm0.020$ & $0.914\pm0.040$ & \boldmath{$0.922\pm0.018$} & $0.910\pm0.029$ \\ [2pt]
        Surprise    & $0.993\pm0.009$ & $0.995\pm0.021$ & \boldmath{$0.995\pm0.008$} & $0.993\pm0.012$ & $0.991\pm0.009$& $0.986\pm0.007$\\ [2pt]
	\bottomrule
\end{tabular}}
\end{table}

\begin{table}[!ht]
	\caption{Classification accuracy for binary R-EMO network.}
	\vspace{-6pt}
	\label{tab:net_res_R}
	\centering
	\footnotesize
	\makebox[1.0\textwidth]{\begin{tabular}{l@{\quad}c@{\quad}c@{\quad}c@{\quad}c@{\quad}c@{\quad}c}
	\toprule
     &  468 & 234 & 128 & 64 & 32 & 16  \\
	\midrule
	Anger       & \boldmath{$0.889\pm0.018$} & $0.874\pm0.032$ & $0.862\pm0.04$ & $0.874\pm0.033$ &  $0.883\pm0.047$ &  $0.866\pm0.043$ \\ [2pt]
	Disgust     & $0.941\pm0.023$ & $0.941\pm0.032$ & \boldmath{$0.953\pm0.019$} & $0.897\pm0.034$ &  $0.905\pm0.023$& $0.887\pm0.048$\\ [2pt]
	Fear        & $0.934\pm0.033$ & \boldmath{$0.948\pm0.016$} & $0.939\pm0.027$ & $0.936\pm0.003$& $0.924\pm0.025$ & $0.907\pm0.032$\\ [2pt]
	Happiness   & $0.968\pm0.022$ & \boldmath{$0.978\pm0.011$} & $0.971\pm0.016$ & $0.957\pm0.021$ &  $0.966\pm0.011$& $0.964\pm0.017$\\ [2pt]
	Sadness     & $0.907\pm0.031$ & $0.893\pm0.021$ & $0.893\pm0.037$ & \boldmath{$0.907\pm0.027$} & $0.898\pm0.033$ & $0.893\pm0.033$\\ [2pt]
        Surprise    & $0.962\pm0.014$ & \boldmath{$0.981\pm0.009$} & $0.976\pm0.0016$ & $0.981\pm0.017$ & $0.974\pm0.027$& $0.974\pm0.021$\\ [2pt]
	\bottomrule
\end{tabular}}
\end{table}
\pagebreak

All of the experiments were conducted utilizing Python programming language and Keras{\textsuperscript{\textregistered}} library with the backend operating on a machine equipped with an AMD Ryzen{\textsuperscript{\texttrademark}} 7 5800X 8-core CPU, clocked at 3.80 GHz, 64 GB of RAM, and an NVIDIA{\textsuperscript{\textregistered}} GeForce{\textsuperscript{\texttrademark}} RTX 3080 Ti GPU, clocked at 1.365 GHz and 12,288 MB of GDDR6X RAM. 
The GPU was utilized for training, testing, as well as for the grid search procedure.

We remark that the main purpose of the proposed paper is the description and the implementation of an innovative explainability framework, also adopting an original procedure for the selection of the input features used to train the binary DNNs for emotion recognition.
For this reason, the scope of this section is the description of the results improvements, which are achieved in terms of accuracy of the classification and explainability of the proposed model when compared with the other ones discussed in \cite{DILUZIO2023104418}.
The explainability framework presented in this paper is employed to select the most important input features for the classification of each of the six typical emotions.
Given these premises, the input combination of features for each of the implemented solutions differs from one emotion to the other and hence, it is not possible to follow the approach described in \cite{DILUZIO2023104418} for the construction of a six-emotions classifier by merging the six different binary deep neural models.
In Sect.~\ref{sect:discussion}, the results will be analyzed, commented and compared with similar neural networks trained on the same dataset.
In particular, the two solutions employed for benchmarking purposes and trained on the same dataset are presented in \cite{kumar2016real}:
\begin{itemize}
    \item the first model is a linear-kernel Support Vector Machine (linear SVM), it achieves the following level of accuracies for the different emotions: Anger $85\%$, Fear $95\%$, Disgust $78\%$, Happiness $97\%$, Sadness $75\%$, Surprise $99\%$;
    \item the second model is an SVM using a radial basis function as a kernel (RBF SVM), it achieves the following level of accuracies for the different emotions: Anger $84\%$, Fear $95\%$, Disgust $74\%$, Happiness $98\%$, Sadness $79\%$, Surprise $100\%$.
\end{itemize}

\section{Discussion}
\label{sect:discussion}
In the following, we discuss the results presented in Sect.~\ref{sect:results}.
First of all, in Fig.~\ref{fig:surprise} it is shown which are the most important landmarks for the recognition of a single emotion. 
It is interesting to note how, by reducing the number of landmarks from $468$ to $234$, the facial points that intuitively are not influenced by emotional feeling are the ones depicting the nose and some points on the facial outline. 
When deleting additional input features based on their importance attribution, the facial points depicting the outline of the eyes and the cheeks are progressively erased, mostly highlighting the lips and their boundary. 
This behavior matches human intuition considering that, particularly for the surprise emotion, the human mouth is the body part that shows the most important change. 

Analyzing the numerical results, it can be seen from Table~\ref{tab:net_res_T} and Table~\ref{tab:net_res_R} that optimizing the number of landmarks leads to interesting results in terms of improvements in the accuracy performance of the binary networks. 
The first obvious advantage is represented by the general increase in the classification accuracy of each network. 
Generally speaking, it can be noted that the most performing combinations are obtained using $234$ and $128$ landmarks. 
In particular, both for T-EMO and R-EMO the optimization of the number of input features results in a higher classification accuracy for each emotion. 
The highest improvement in the T-EMO performance is related to the Disgust and Fear emotions, for which there is an increase in the accuracy of $3\%$.
On the other hand, the improvement achieved by the randomized fashion of the network (i.e., R-EMO) is $1.9\%$ for the Surprise and $1.5\%$ for the Fear classification accuracy.

Besides the numerical aspects, the proposed methodology allows the construction of explainable DNNs that are also more accurate.
Moreover, when comparing the accuracy achieved with the proposed method with the selected benchmarks, the T-EMO network achieves higher classification accuracy for the entire set of emotions when compared with the linear SVM and for five emotions out of six when compared with the RBF SVM.
The R-EMO architecture, on the other hand, achieves higher classification accuracy for five emotions with respect to the Linear SVM and for four emotions out of six when compared with the RBF SVM.
The average increase in the accuracy achieved by T-EMO with the two benchmarking models is $8\%$ with respect to the linear SVM and $9\%$ when compared with the RBF SVM.
Considering the R-EMO network, the average increase in the accuracy is $6\%$ with respect to the Linear SVM and $7\%$ when compared with the RBF SVM.

Something that must be highlighted is that R-EMO accuracy performance in its best landmarks configuration is lower than the relative T-EMO accuracy for every emotion. 
This is interesting because it is something that does not occur with the basic $468$ facial point configuration. 
However, this approach demonstrates that exploring explainable randomized architectures is a promising concept to investigate. This is because the decrease in accuracy caused by randomizing one or more layers can be offset not only by a significant reduction in training time, but also by an additional improvement in both efficiency and effectiveness performance through explainability-based feature optimization.

The general increase in the network accuracy, caused by a diminution in the number of input features to the network, represents a great advantage consisting of the overall improvement of the network classification performance whit a substantial decrease in the needed computational power. 
In fact, by this method, the input feature space is controlled by a mathematically based explanation evaluating the most informative input features after training a deep binary neural classifier. 
The numerical results are even more impressive when the improvement margin for each of the binary networks implemented is at most $10\%$. 
When further diminishing the number of input landmarks, the model is still able to obtain an accurate prediction of the video class, but the accuracy generally decreases visibly when using less than 64 landmarks.

To conclude the discussion, emotion recognition from video analysis is inherently complex due to the dynamic nature of facial expressions, to the general absence of a common structure for the facial landmarks, and to the possibility of simulating a different emotion from the one really experienced by a person in a determined moment. 
The scope of this investigation, which is also one of the main novelties of the implemented methodology, relies on the attempt to find common patterns describing a general behavior that is representative of each typical emotion. 
Even if this is a challenging task for a neural network, the explainability framework presented in this paper seems to be an effective method for the classification of the most important facial landmarks, meant as the ones that contribute the most to the optimization of the neural network parameters for the classification of the single emotions.

\section{Conclusions}
\label{sect:conclusions}
Given the continuous interest in DL techniques and model functioning in the biomedical context, it is of remarkable importance to find new efficient explainability solutions to exploit the inner mechanism dictating the behavior of neural networks in this domain. 
From a different point of view, it is of great interest to investigate the trade-off between the accuracy of the classification and the computational cost of the optimization procedure; hence, the merging of explainability and randomization of some layers of the network is particularly intriguing from a scientific point of view. 
In this paper, we applied a customized fashion of the explainability technique of IGs to binary DNNs with randomized convolutional layers to capture the functioning of the network optimizing the number and position of input features. 
In particular, the networks were trained on videos of emotional feeling, from which $468$ facial landmarks were extracted and whose coordinates are employed as input.

The aim of this work was to optimize the number of landmarks to use as input features for each emotion evaluating the diminishing of the noise feeding the model and the consecutive increase in the overall classification performance. 
Investigating the obtained attribution of importance for each input feature, we can confirm, even from a graphical point of view, that the implementation of the explainability method presented in this paper rests on solid foundations. The obtained results show that the landmarks selected as more important are the ones situated in the parts of the face that change the most during the emotional feeling, they are not always the same for the six typical emotions. 
On the other hand, by analyzing the numerical results obtained with the optimized networks, we can confirm the edge of the proposed approach, reporting advantages in classification accuracy and computational cost needed when using the most important selected landmarks for the training procedure.

In order to assess the randomized nature of the proposed method, we employed the fully trained version of the same network and the model presented in \cite{DILUZIO2023104418}, optimized using the whole set of $468$ face landmarks as input to the neural networks. 
Given the improvements obtained in this work with this customized version of IGs, this method can pave the way for even more complex investigations on the optimization of input features to DNNs for emotion recognition, exploiting even more in-depth, the inner mechanism ruling the functioning of deep neural architectures. 
Moreover, this method can be applied to each similarly-implemented binary network, showing high portability which could result helpful in several different real-world applications and practical fields like surveillance systems, security systems in general, and even amusement tools.

From a practical point of view, given its mathematical implementation, the method proposed herein generates an attribution mask that reflects the facial landmarks movement during the emotional feeling.
In detail, this attribution mask is representative of the importance of each landmark in the videos during the optimization procedure of the DNN weights. 
As changing the input parameters for analyzing the results can lead to different outcomes, a particularly important novel aspect of the proposed methodology relies on its possible application to several models fed with different data sources.
In particular, it is possible to perform the optimization of the input frame described in this paper when the computation of the integrated gradients is feasible, independently of the input data and its dimensionality.

Finally, the methodology presented in this paper can pave the way for several different studies on new explainability techniques that can be used to describe the inner mechanism of deep learning models in several real-world application fields. 
In the future, the possibility to increase the global explanation capability of the proposed method can be achieved with the employment of more emotion recognition datasets, where other combinations of input features can be investigated and newer and/or deeper architectures can be analyzed.
In this sense, the methodology presented in this paper could be also improved with an innovative implementation suited for multimodal data and architectures, in order to study the dependencies and the relevance of the adopted approach during the optimization procedure.

\label{sect:references}

\end{document}